\documentclass[a4paper,UKenglish,cleveref, autoref, thm-restate]{lipics-v2021}

\makeatletter
\@ifundefined{iflatexml}{}{%
  \newcommand{\hideLIPIcs}{} 
  \renewcommand{\ccsdesc}[2][]{} 
  \renewcommand{\relatedversion}[1]{}
  \renewcommand{\category}[1]{}
  \renewcommand{\Copyright}[1]{}
}
\makeatother

\pdfoutput=1 
\hideLIPIcs  

\usepackage{booktabs}
\usepackage{float}

\usepackage{hyperref}

\title{Transfer Learning from Foundational Optimization Embeddings to Unsupervised SAT Representations} 

\titlerunning{Transfer Learning from Foundational Optimization Embeddings to SAT} 

\makeatletter
\@ifundefined{iflatexml}{
  \author{Koyena {Pal}}{AI Center of Excellence, Fidelity Investments, USA\and Khoury College of Computer Sciences, Northeastern University, USA}{pal.k@northeastern.edu}{}{}
  \author{Serdar Kad\i o\u{g}lu}{AI Center of Excellence, Fidelity Investments, USA\and Department of Computer Science, Brown University, USA}{serdark@cs.brown.edu}{}{}
  
  \authorrunning{Koyena Pal and Serdar Kad\i o\u{g}lu}
}{
  \author{Koyena Pal\textsuperscript{1,2} \quad Serdar Kadıoğlu\textsuperscript{1,3} \\[1em]
    \parbox{\textwidth}{\centering \small
      \textsuperscript{1}AI Center of Excellence, Fidelity Investments, USA \\
      \textsuperscript{2}Khoury College of Computer Sciences, Northeastern University, USA \\
      \textsuperscript{3}Department of Computer Science, Brown University, USA \\
      \vspace{0.5em}
      \texttt{pal.k@northeastern.edu, serdark@cs.brown.edu}
    }
  }
  \date{} 
}
\makeatother

\Copyright{CC-BY} 



\category{} 

\relatedversion{} 

\nolinenumbers 

\begin{document}

\maketitle

\begin{abstract}
Foundational optimization embeddings have recently emerged as powerful pre‑trained representations for mixed‑integer programming (MIP) problems. These embeddings were shown to enable cross‑domain transfer and reduce reliance on solver‑generated labels. In this work, we investigate whether such representations generalize beyond optimization to decision problems, focusing on Boolean satisfiability (SAT). We adapt the foundational optimization architecture to SAT by mapping CNF formulas into the same bipartite constraint–variable graph representation used for MIPs. This allows direct reuse of the pre‑trained embedding model without architectural changes or supervised fine‑tuning. Our results show that these embeddings capture structural regularities in SAT instances and support unsupervised tasks such as instance clustering and distribution identification. We demonstrate, for the first time, that foundational optimization embeddings can transfer to constraint satisfaction domains. Our findings is a step toward a unified representational framework for both optimization and decision problems.
\end{abstract}


\vspace{-0.12cm}
\section{Introduction}
\label{sec:introduction}
\vspace{-0.15cm}
Mixed-integer programming (MIP) and Boolean satisfiability (SAT) are central formalisms for combinatorial reasoning for a wide range of optimization and decision problems. Recent work has shown that graph neural networks (GNNs) can learn powerful structural representations for specific problem families. Unfortunately, such models typically rely on solver-generated labels, must be retrained for each new domain or task, and often generalize poorly across distributions and instance scales.

\smallskip
Foundational optimization embeddings, such as \textsc{Forge}~\cite{forge}, offer an alternative direction. They learn generic representations of MIP instances via \textit{unsupervised pretraining without dependency on labels} on large, heterogeneous corpora. It has been shown to support transfer across optimization tasks such as cut generation and search guidance. This raises a natural question: \textit{\textbf{can a model designed for and trained on purely for MIP structure provide useful representations for decision problems such as SAT, and if so, how far does this cross-domain transfer extend?}}

\smallskip
In this paper, we investigate transfer learning from foundational optimization embeddings to SAT. We consider several variants of the original \textsc{Forge} architecture~\cite{forge} that span different degrees of transfer. In \textsc{Forge-Mip}, we show how to directly reuse the original \textsc{Forge} model and its MIP-specific features on SAT instances encoded as MIPs. In \textsc{Forge-Mip-Sat}, we keep the pretrained weights and architecture but replace MIP node features with SAT-specific clause and variable features. Finally, in \textsc{Forge-Sat}, we retain the architecture but discard the MIP weights and pretrain from scratch on SAT instances using SAT features. 

This work provides an initial empirical study of these variants on unsupervised tasks. Our main contributions are:
\begin{enumerate}
    \item \textbf{Architecture transfer to SAT}: We adapt the \textsc{Forge} foundational architecture, originally developed for MIPs, to SAT via a SAT-to-MIP encoding pipeline and SAT-specific node features.
    \item \textbf{Weight-level transfer from MIP to SAT:} We show that pretrained MIP embeddings, \textsc{Forge-Mip}, already capture useful SAT structure, and that feature specialization, \textsc{Forge-Mip-Sat} further improves performance.
    \item \textbf{SAT-native foundational model:} We introduce \textsc{Forge-Sat}, a SAT-pretrained variant that reuses the Forge training paradigm while specializing to SAT instances.
    \item \textbf{Empirical evaluation on SAT benchmarks:} We evaluate all variants on unsupervised clustering. We provide evidence for cross-domain generalization and outlining the limits and opportunities of foundational embeddings across optimization and decision problems. 
    \item \textbf{Open‑source pretrained models and pipelines:} All pretrained models and training pipelines are released open-source. This enables generating meaningful embeddings for arbitrary SAT instances, clauses, and variables \textit{out-of-the-box}, as in the success of pretrained text, image, and audio embeddings.

\end{enumerate}

\section{Background}
\label{sec:background}
We briefly review the SAT and MIP formalisms, their bipartite graph representations, and outline the standard transformation that encodes a SAT formula as a MIP instance.

\medskip
\noindent \textit{Boolean Satisfiability (SAT):} A propositional formula in conjunctive normal form (CNF) is a conjunction of clauses
\(F = C_1 \land \cdots \land C_m\), where each clause \(C_j\) is a disjunction of literals over Boolean variables \(x_1,\dots,x_n\). A literal is either a variable \(x_i\) or its negation \(\lnot x_i\).
The SAT problem asks whether there exists an assignment \(\mathbf{x} \in \{0,1\}^n\) such that all clauses are satisfied.

\smallskip
\noindent \textit{Mixed-Integer Programming (MIP):} A mixed-integer program consists of linear constraints over continuous and integer variables: \(\min_{\mathbf{x}} c^\top \mathbf{x} \;\text{s.t.}\; A\mathbf{x} \le b,\;
\mathbf{x}_{\mathcal{I}} \in \mathbb{Z}^{|\mathcal{I}|},\; \mathbf{x}_{\mathcal{C}} \in \mathbb{R}^{|\mathcal{C}|}\).
where \(A \in \mathbb{R}^{m \times n}\), \(b \in \mathbb{R}^m\), and the variable set is partitioned into
integer variables \(\mathcal{I}\) and continuous variables \(\mathcal{C}\).

\smallskip
\noindent \textit{MIP Bipartite Graph:} Following~\cite{gasse2019exact}, it is common to represent a MIP instance as a bipartite graph with constraint nodes and variable nodes, and an edge between constraint \(i\) and variable \(j\) whenever \(A_{ij} \neq 0\). In \textsc{Forge}~\cite{forge}, node features encode properties of constraints (e.g., sense, right-hand side) and variables (e.g., type, bounds, objective coefficient).

\smallskip
\noindent \textit{Encoding SAT as MIP:} A CNF formula can be encoded as a 0–1 MIP by introducing a binary variable \(x_i\) for each Boolean variable and a linear inequality for each clause. For a clause \(C_j = (\ell_{j1} \lor \cdots \lor \ell_{jk})\), we define
\(
\sum_{t=1}^k y_{jt} \;\ge\; 1,
\)
where each literal \(\ell_{jt}\) is mapped to a term
\(y_{jt} = x_i\) if \(\ell_{jt} = x_i\), and
\(y_{jt} = 1 - x_i\) if \(\ell_{jt} = \lnot x_i\).
The resulting system of linear constraints over binary variables is satisfiable if and only if the original CNF formula is satisfiable. 

\smallskip
This SAT-to-MIP encoding allows us to reuse the bipartite constraint–variable graph representation and the foundational \textsc{Forge} architecture for SAT instances.

\begin{figure}[t]
    \centering
    \includegraphics[width=\linewidth]{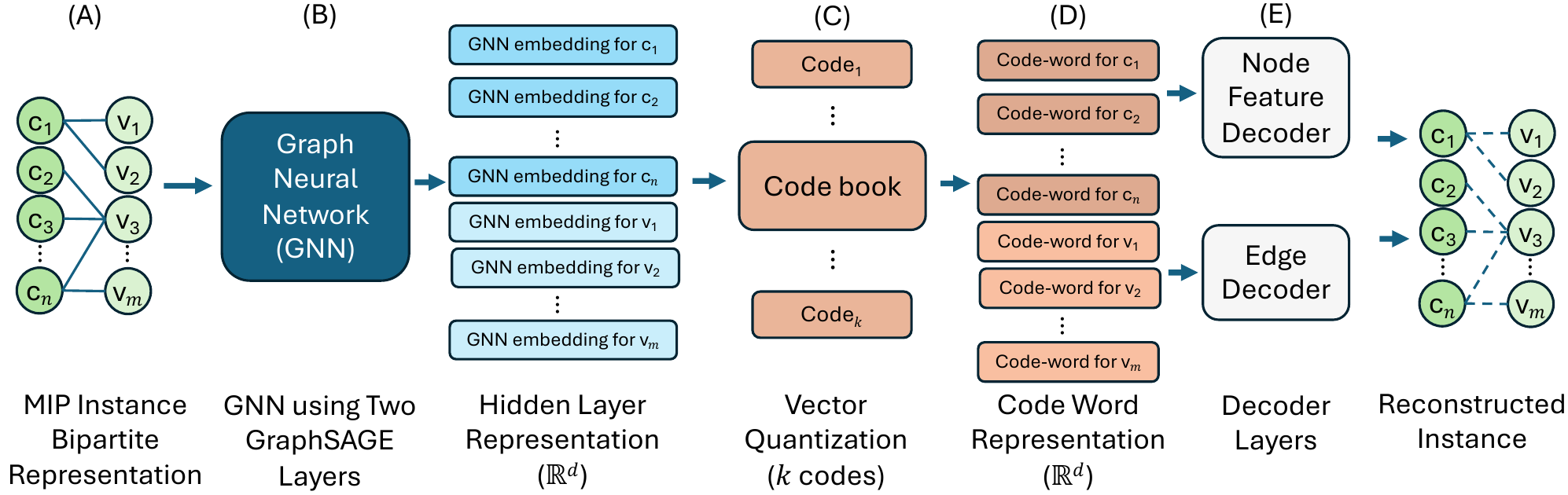}
    \caption{Overview of \textsc{Forge} architecture~\cite{forge}: Starting with the bipartite representation of MIPs and its GNN embedding (A-B), \textsc{Forge} uses vector quantized graph autoencoder (C-D) to reconstruct node features and edges (E). This enables learning generic MIP representations with a pre-trained model. 
    }
    \label{fig:forge-overview}
\end{figure}

\section{Methodology}
\label{sec:methodology}
\subsection{Foundational Architecture}
\textsc{Forge} is a foundational representation-learning framework that learns generic embeddings of mixed-integer programs (MIPs) through \textit{unsupervised pretraining} on large, heterogeneous corpora. 

\smallskip
As illustrated in Figure~\ref{fig:forge-overview}, a MIP instance is represented as a bipartite graph of constraint and variable nodes, each equipped with MIP related features. A two-layer GraphSAGE~\cite{hamilton2017inductive} encoder produces latent node embeddings, which are discretized through a vector-quantization (VQ) codebook~\cite{yangvqgraph}. A decoder reconstructs
node features and incident edges from these codewords. The training objective combines node-feature reconstruction, edge reconstruction, and VQ commitment losses. 

\smallskip
Once pretrained over a diverse MIP corpora, \textsc{Forge} provides:

\begin{itemize}
    \item \textbf{Instance embeddings:} the empirical distribution of codewords assigned to all nodes in an instance, serving as a global representation.
    \item \textbf{Constraint \& variable embeddings:} codewords associated with individual constraint and variable nodes provides embeddings for nodes.
\end{itemize}

The \textsc{Forge} architecture is \textit{entirely unsupervised and does not rely on solver labels}, making it a natural candidate for cross-domain transfer.





\subsection{FORGE Applied to SAT}
To apply \textsc{Forge} to SAT, we first encode each CNF formula as a MIP using the SAT-to-MIP transformation described in Section~\ref{sec:background}. The resulting bipartite graph has the same structure expected by \textsc{Forge}, allowing us to reuse the architecture directly. We explore three variants that differ in how features and pretrained weights are used.



\begin{figure}[t]
    \centering
    \includegraphics[width=\linewidth]{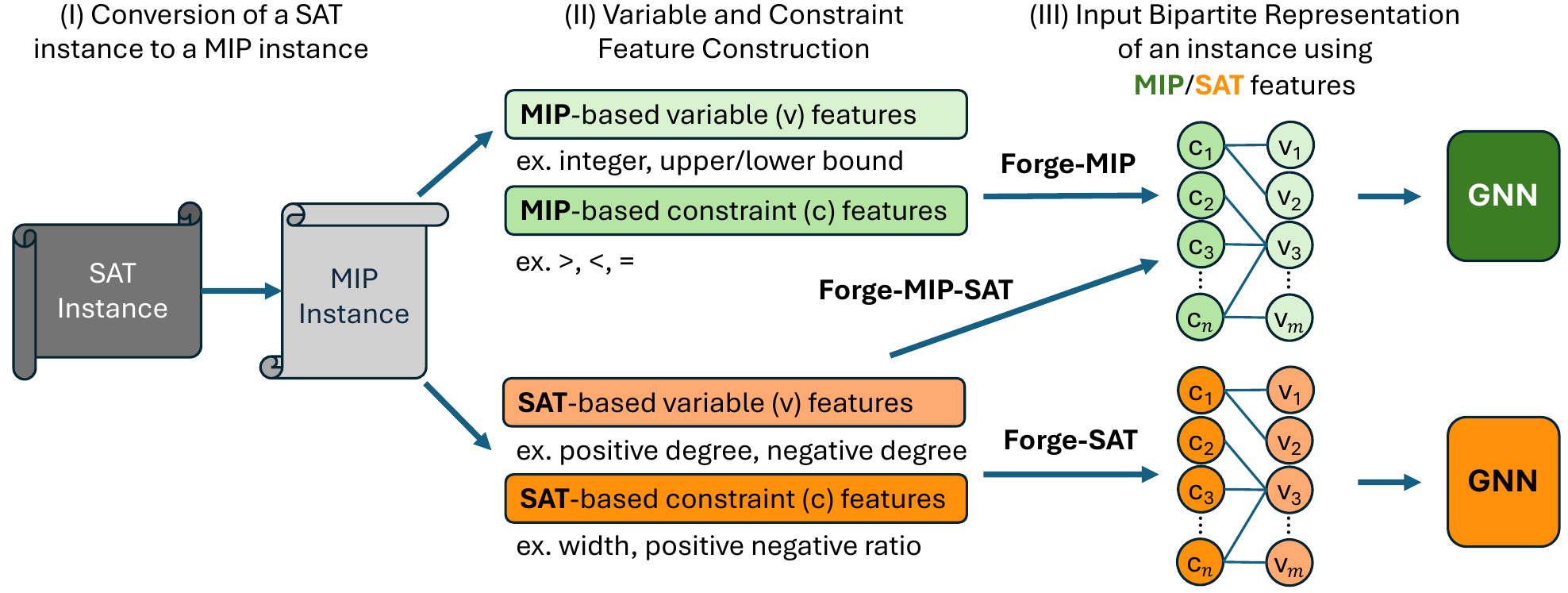}
    \caption{Our transfer learning setup for \textsc{Forge} applied to SAT to enable unsupervised SAT representations. (I) convert SAT instances into MIP form. (II) construct either MIP or SAT node feature. (III) bipartite representation, with MIP or SAT node features, are fed into \textsc{Forge} architecture trained on MIP or SAT instances. Overall, we obtain SAT representations (embeddings) at the instance-, clause-, and variable level.}
    \label{fig:method-overview}
\end{figure}

\medskip
\noindent \textbf{FORGE-MIP (zero‑change transfer):} This variant, as shown in~\cref{fig:method-overview} (upper side), directly applies the pretrained \textsc{Forge} model to SAT instances encoded as MIPs. Both the architecture and the MIP-specific node features are kept unchanged. This
represents pure weight-level transfer from optimization to satisfiability.

\medskip
\noindent \textbf{FORGE‑MIP-SAT (feature‑adapted transfer):}
\label{sec:forge-mip-sat}
This variant also reuses the pretrained MIP weights but replaces the MIP node features with SAT-specific features. SAT-specific constraint node features are described by clause-level properties: \texttt{width} (number of literals in the clause), \texttt{pos$\_$count} (number of positive literals), \texttt{neg$\_$count} (number of negative literals), and \texttt{pos$\_$neg$\_$ratio} (ratio of positive to negative literals). SAT-specific variable node features are described by degree-based properties: \texttt{degree} (total number of clauses the variable appears in), \texttt{pos$\_$deg} (number of clauses where it appears positively), \texttt{neg$\_$deg} (number of clauses where it appears negated), \texttt{pos$\_$neg$\_$ratio} (ratio of positive to negative appearances), \texttt{pos$\_$deg$\_$norm} (positive degree normalized by the mean positive degree across all variables), and \texttt{neg$\_$deg$\_$norm} (negative degree normalized by the mean negative degree across all variables). The graph structure and pretrained weights remain unchanged, isolating the effect of feature specialization.



\smallskip
\noindent \textbf{FORGE‑SAT (SAT‑native model):}
This variant, as shown in~\cref{fig:method-overview} (lower side), retains the \textsc{Forge} architecture but discards the MIP-pretrained weights. Using the SAT-specific features above, we pretrain the full VQ-GNN model from scratch on SAT instances from the G4SATBench training split. This represents architecture-level transfer: the foundational training paradigm is reused, but the representation is learned entirely from SAT data.


\smallskip
Across these variants, we study two axes of transfer learning:
\begin{itemize}
    \item \textbf{Weight-level transfer:} whether MIP-pretrained parameters provide useful structure for SAT, evaluated via \textsc{Forge-Mip} and \textsc{Forge-Mip-Sat}.
    \item \textbf{Architecture-level transfer:} whether the \textsc{Forge} training paradigm generalizes to SAT when trained from scratch via \textsc{Forge-Sat}.
\end{itemize}




\section{Computational Experiments}
\label{experiments}
Our experiments evaluate whether foundational optimization embeddings transfer effectively from MIP to SAT and whether unsupervised pretraining yields representations that capture structural regularities across diverse SAT distributions.

For that purpose, we study two axes of variation: (1) the source of pretraining, comparing models trained on MIP instances versus SAT-induced instances, and (2) generalization to unseen SAT domains. Our preliminary results focus on unsupervised evaluation (\cref{sec:unsupervised}) based on clustering instance-level embeddings of unseen formulas. This analysis assesses (i) how well the learned representations differentiate SAT problem families, and (ii), separate satisfiable from unsatisfiable instances.



\vspace{-0.2cm}
\section{Unsupervised Clustering}
\label{sec:unsupervised}
In these experiments, we cluster the instance-level embeddings of \emph{unseen} SAT formulas using different embedding variants and evaluate how well the resulting clusters recover the underlying problem distributions and SAT/UNSAT categories.

\vspace{-0.2cm}
\subsection{G4SATBench Dataset}
\label{sec:unsupervised-datasets}
We use SAT instances from G4SATBench~\cite{li2024gsatbench}, a widely used benchmark for GNN-based SAT reasoning. The benchmark contains seven datasets drawn from three domains: \emph{random}, \emph{pseudo-industrial}, and \emph{combinatorial} SAT problems. The random category includes instances generated by the SR model~\cite{selsam2018learning} and the 3-SAT generator in CNFGen~\cite{lauria2017cnfgen}. The pseudo-industrial category includes Community Attachment (CA)~\cite{giraldez2015modularity} and Popularity–Similarity (PS) 
instances~\cite{giraldez2017locality}, which mimic structural properties of real-world SAT formulas. The combinatorial category includes $k$-Clique, $k$-Dominating Set, and $k$-Vertex Cover instances, also generated via CNFGen~\cite{lauria2017cnfgen}. 

\smallskip
Each dataset provides easy, medium, and hard difficulty levels with dedicated train and test splits, and each split is balanced between SAT and UNSAT instances. This diversity makes G4SATBench a suitable testbed for evaluating cross-domain generalization of SAT embeddings.

\smallskip
In our clustering experiments, we use all test instances from the Hard category, yielding 1400 \emph{unseen} formulas: 200 from each of the seven problem types, with an equal split of 100 SAT and 100 UNSAT instances. This results in 14 ground-truth groups defined by (problem type, feasibility) pairs. We focus on the Hard split for clarity, as results on the Easy and Medium categories follow similar trends.




\vspace{-0.2cm}
\subsection{Comparisons} 
We compare the clustering quality obtained from the instance-level embeddings produced by our three transfer-learning variants introduced in Section~\ref{sec:methodology}: \textsc{Forge-Mip}, \textsc{Forge-Mip-Sat}, and \textsc{Forge-Sat}. The first two variants reuse the pretrained \textsc{Forge}~\cite{forge} weights learned on MIP instances, while
\textsc{Forge-SAT} is pretrained from scratch on the G4SATBench training set, which contains 28,781 instances.

\medskip
\noindent\textit{Control Baseline:} As a baseline, we include a simple static heuristic, \textsc{Static-Sat}, which uses the same SAT-specific clause and variable features described in~\cref{sec:forge-mip-sat}. Clause and variable features are mean-pooled at the instance level and concatenated to form a fixed, non-learned instance embedding.

\medskip
\noindent\textit{Evaluation Metrics:} We evaluate clustering quality using
\emph{Normalized Mutual Information (NMI)} and \emph{Purity}. NMI measures the agreement between predicted clusters and the 14 ground-truth categories, ranging from 0 to 1, with higher values indicating stronger alignment. Purity quantifies how homogeneous each cluster is by computing the fraction of instances belonging to its dominant category and averaging across clusters. Similar to NMI, Purity ranges from 0 to 1, with higher values indicating more coherent clusters.




\subsection{Clustering Results}
\begin{figure}[t]
    \centering
    \includegraphics[width=\linewidth]{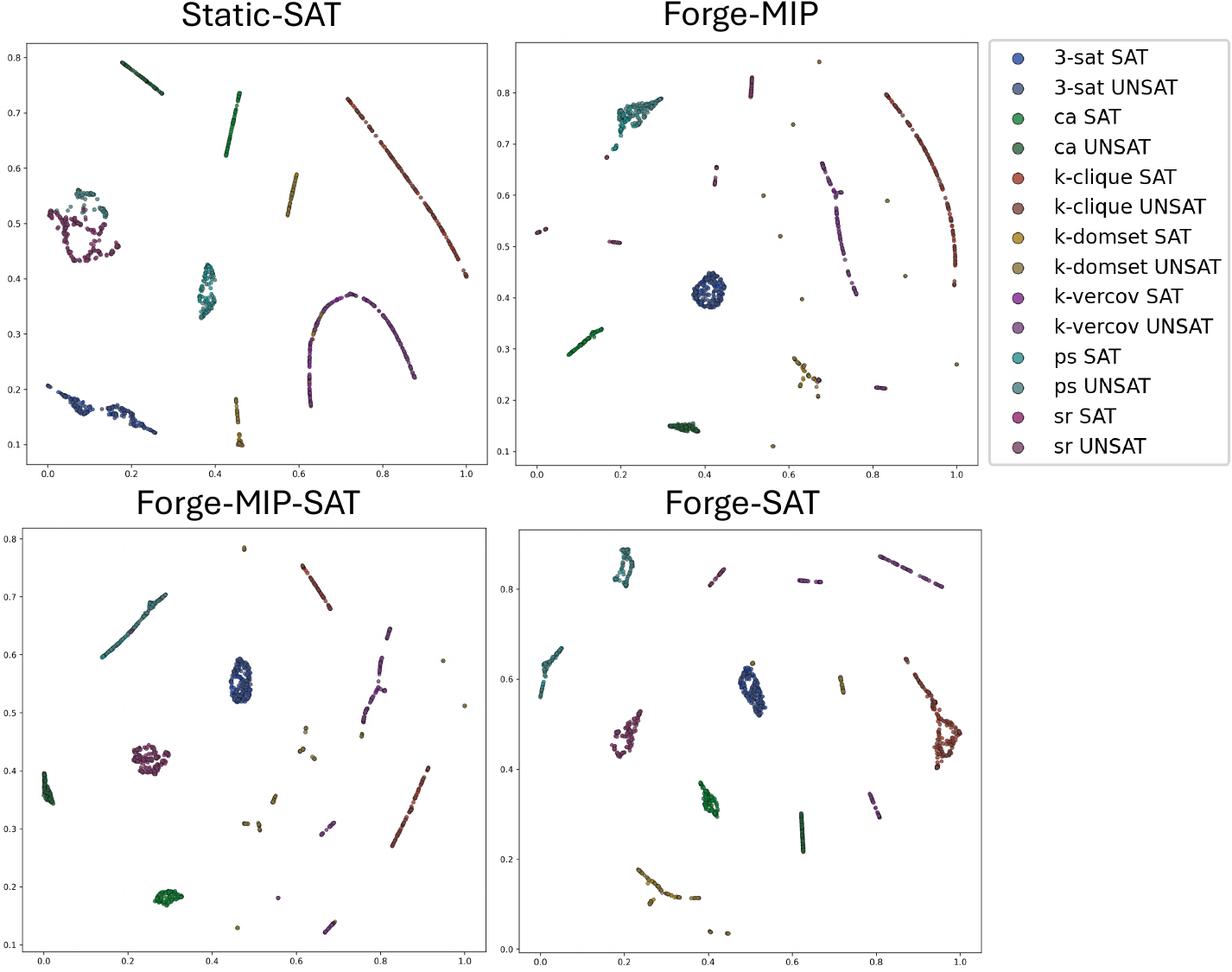}
    \caption{Clustering SAT embeddings across static and different transfer learning featurizations.}
    \label{fig:hard-clusters}
\end{figure}

Figure~\ref{fig:hard-clusters} shows the instance-level embeddings projected into 2D using PaCMAP~\cite{wang2021understanding}. In an ideal scenario, the embeddings would form 14 distinct clusters corresponding to the seven problem types and their SAT/UNSAT labels. \textsc{Forge-SAT} comes closest to this ideal, \textbf{producing 13 compact and well-separated clusters}; the only exception is that $k$-Clique SAT and UNSAT instances overlap (see on the right side of \textsc{Forge-Sat} sub-plot). Notably, \textsc{Forge-SAT} also cleanly separates feasibility within the same problem type, see, e.g., PS SAT vs.\ UNSAT on top-left of  \textsc{Forge-Sat} sub-plot.

\smallskip
In contrast, \textsc{Static-Sat} recovers only nine clusters, with substantial mixing across problem types, and both \textsc{Forge-MIP} and \textsc{Forge-MIP-SAT} yield less compact and less clearly separated clusters than \textsc{Forge-SAT}. This difference reflects the fact that \textsc{Forge-MIP} and \textsc{Forge-MIP-SAT} rely on weights pretrained on \emph{optimization instances from MIPLIB}, whereas \textsc{Forge-SAT} is pretrained directly on SAT instances from G4SATBench.


\smallskip
Table~\ref{tab:hard_cluster_scores} reports the quantitative clustering results. For NMI, a perfect score is 1.0, whereas a uniform random assignment over the 14 ground‑truth groups would yield roughly $1/14 \approx 0.07$. 

\smallskip
First, we observe that static SAT features already provide a strong baseline, improving NMI from the random baseline of $0.07$ to $0.74$. Both \emph{pretrained optimization embeddings}, \textsc{Forge-MIP} and \textsc{Forge-MIP-SAT}, further increase NMI, demonstrating that representations learned from MIP data transfer meaningfully to SAT. It is remarkable that embeddings trained on the \textsc{MIPLIB} dataset transfer so effectively to the G4SATBench dataset. Among the two, \textsc{Forge-MIP-SAT} performs best: using the same pretrained weights as \textsc{Forge-MIP} but replacing MIP node features with SAT-specific node features yields gains in both NMI and Purity.

\smallskip
These results further improve when training directly on SAT data: \textsc{Forge-SAT} achieves an \textbf{NMI of 0.79 and a Purity of 0.66}, demonstrating that the \textsc{Forge} architecture and hyperparameters transfer effectively to the SAT domain \textit{without any modification}. 


\smallskip
Note that here we are not tuning any hyperparameters of \textsc{Forge}; all models use the original optimization‑tuned settings. Conducting SAT‑specific pretraining and hyperparameter tuning may further improve performance. Nevertheless, all three \textsc{Forge} variants already provide a practical method for obtaining meaningful SAT embeddings directly \emph{out-of-the-box}.

\smallskip
To the best of our knowledge, this is the first approach that transfers a foundational optimization model to SAT and produces instance‑level SAT embeddings in a fully unsupervised manner using a pretrained model. This capability allows obtaining meaningful SAT representations without any SAT‑specific supervision or dependency on SAT solver and constitutes the primary contribution of our work.

\begin{table}[t]
\renewcommand{\arraystretch}{1.2}
\centering
\begin{tabular}{lcccc}
\toprule
Metric & \textsc{Static-Sat} & \textsc{Forge-Mip} & \textsc{Forge-Mip-Sat} & \textsc{Forge-Sat} \\
\midrule
NMI    & $0.74 \pm 0.002$ & $0.76 \pm 0.001$ & $0.77 \pm 0.003$ & \textbf{0.79 $\pm$ 0.016} \\
Purity & $0.63 \pm 0.002$ & $0.61\pm 0.002$ & \textbf{0.66 $\pm$ 0.001} &  \textbf{0.66 $\pm $0.002} \\
\bottomrule
\end{tabular}
\vspace{0.2cm}
\caption{NMI and Purity scores across methods on Hard G4SATBench test instances.}
\label{tab:hard_cluster_scores}
\end{table}

\section{Discussion and Limitations}
Our findings show that \textsc{Forge-Sat}, while retaining the original optimization-oriented \textsc{Forge} architecture, achieves promising results. These findings suggest that the foundational training paradigm
transfers beyond optimization and can learn meaningful SAT structure directly.

\smallskip
Our initial study can be improved in several directions. First, and most importantly, beyond the unsupervised setting, we can showcase the usefulness of the generated embeddings in downstream tasks such as satisfiability prediction, assignment prediction, core UNSAT prediction, or solver-guided heuristics. Second, our SAT node features are simple statistical descriptors rather than carefully engineered or learned representations, and ablation studies may reveal more effective feature subsets. Third, our pretraining corpus is limited to a single SAT benchmark; larger and more diverse collections may yield stronger foundational embeddings without any changes to the overall architecture.

\smallskip
Looking forward, we aim to scale \textsc{Forge-Sat} pretraining, study supervised prediction tasks, explore integration with SAT solvers, and extend this framework to broader classes of constraint satisfaction problems





\section{Related Work}
\label{sec:related}

Graph neural networks have become the dominant paradigm for learning representations of SAT instances. NeuroSAT~\cite{selsam2018learning} demonstrated that a message-passing network trained to predict satisfiability can implicitly learn to search for satisfying assignments, spurring a broad line of work on GNN-based SAT solvers. One direction integrates neural predictions into solver pipelines: NLocalSAT~\cite{ijcai2020p164} uses network predictions to initialize variable assignments for local search solvers, NeuroGlue~\cite{han2020enhancingsatsolversglue} predicts variables in glue clauses to guide Glucose-family solvers~\cite{glucose}, and Graph-Q-SAT~\cite{kurin2019improving} formulates variable branching as a Q-learning problem over graph-structured SAT representations. To avoid the cost of online neural inference during solving, a second direction performs neural computation offline: NeuroCore~\cite{selsam2019neurocore} predicts variables likely to appear in unsatisfiable cores during CDCL~\cite{marques2009conflict} solving, NeuroBack~\cite{wang2024neuroback} predicts backbone variables to guide phase selection in Kissat without runtime GPU overhead, and Chen et al.~\cite{10.1145/3716368.3735251} guide variable branching entirely at preprocessing time. \textit{Across both directions, representations are learned in a supervised, task-specific manner tied to a particular solver or objective.}

A parallel line of work has explored more general instance level representations of SAT. Cameron et al.~\cite{cameron2020predicting} encode CNF formulas as permutation-invariant sparse matrices and show that end-to-end learning achieves competitive satisfiability prediction. Duan et al.~\cite{duan2022augment} apply contrastive pretraining to GNN representations of CNF formulas, using CDCL-augmented instances as positive pairs to embed solver-relevant structure into the latent space. For solver selection, GraSS~\cite{zhang2024grass} learns supervised embeddings from literal clause graphs augmented with expert-designed features to predict which solver in a portfolio will perform best on a given instance. Cardillo et al.~\cite{cardillo2025milpsatgnnneuralsatsolver} take a complementary approach by encoding $k$-CNF formulae as MILP-based weighted bipartite graphs and training a GNN to predict satisfiability. They establish permutation and equivalence invariance of the representation, identify a theoretical limitation for foldable formulae where standard GNNs cannot distinguish SAT from UNSAT instances, and prove a universal approximation result under random node initialization, demonstrating that MILP bipartite graph representations are a viable alternative to the literal clause graphs used in NeuroSAT-style approaches.

\textit{Despite this breadth of existing work, prior approaches learn representations only in service of a specific task or solver.} Our work takes a different direction: we ask whether \textit{unsupervised structural embeddings}, pretrained on MIP or SAT instances without any solver feedback, can capture meaningful regularities across SAT domains and feasibility, and whether models pretrained on MIP can transfer to SAT through feature adaptation or domain‑specific pretraining. Our initial results indicate that they can, which opens several promising directions for future work.

\section{Conclusion}
\label{sec:conclusion}
In this paper, we show the possibility of unified structural pretrained models across decision (SAT) and optimization (MIP) problems. It would be interesting to explore how these ideas extend to Constraint Satisfaction Problems (CSPs) and Constraint Programming (CP). It remains an open question whether a single foundational pretrained model can serve all of these paradigms. However, our results show that it is now possible to train a \textsc{Forge} model on a \textit{mixture of MIP and SAT instances, enabling hybrid, multi‑domain pretraining}. This opens an exciting direction toward foundational models that unify optimization and satisfaction and support both supervised and unsupervised tasks.


\bibliography{references}

\end{document}